\documentclass[10pt,twocolumn,letterpaper]{article}

\usepackage[pagenumbers]{cvpr} 

\usepackage{booktabs}
\usepackage{multirow}
\usepackage{adjustbox}
\usepackage{float}
\usepackage{listings}

\usepackage[most]{tcolorbox}

\newtcblisting{promptbox}{
    listing only,
    breakable,
    colback=black!2,
    colframe=black!18,
    boxrule=0.4pt,
    arc=1mm,
    left=6pt,
    right=6pt,
    top=5pt,
    bottom=5pt,
    before skip=4pt,
    after skip=6pt,
    listing options={
        basicstyle=\ttfamily\footnotesize,
        breaklines=true,
        breakatwhitespace=true,
        columns=fullflexible,
        keepspaces=true,
        showstringspaces=false
    }
}

\definecolor{cvprblue}{rgb}{0.21,0.49,0.74}
\usepackage[pagebackref,breaklinks,colorlinks,allcolors=cvprblue]{hyperref}

\def\confName{CVPR}
\def\confYear{2026}

\title{TraceKV: Policy-Preserving Low-Rank KV Cache Compression for Reasoning Vision-Language Models}

\author{
    Ke Li$^{1,*}$,\qquad
    Zishan Shao$^{1}$,\qquad
    Yiran Chen$^{1,\dagger}$\\
    $^{1}$Department of Electrical and Computer Engineering,
    Duke University\\
}

\begin{document}
\maketitle

\begingroup
\renewcommand{\thefootnote}{\fnsymbol{footnote}}
\footnotetext[1]{Work done during an internship at Duke University.}
\footnotetext[2]{Corresponding author.}
\endgroup

\begin{abstract}
Reasoning vision-language models (VLMs) impose substantial KV cache demands from both large multimodal contexts and growing reasoning traces. Low-rank KV compression has emerged as a promising paradigm that stores low-dimensional latents of KV states and can be combined with quantization. However, existing low-rank methods often induce growing predictive policy drift from the dense model when compressing reasoning KV states, resulting in reasoning failures. We identify two key causes of this degradation: calibration-state mismatch between static contexts and on-policy reasoning states, and unequal error sensitivity across KV states. To address these challenges, we propose TraceKV, a training-free low-rank KV compression framework that unifies on-policy calibration, sensitivity-aware SVD, and global rank allocation under a single policy-preserving principle, substantially reducing policy-KL divergence during reasoning. 
Experiments across multiple VLMs and visual reasoning benchmarks demonstrate superior accuracy--memory trade-offs,
achieving near-lossless reasoning accuracy with 5.64$\times$ KV cache compression when combined with quantization. Optimized low-rank attention kernels further deliver up to 1.83$\times$ end-to-end decoding speedup.
Code will be available at https://github.com/AK-DREAM/TraceKV.
\end{abstract}    
\section{Introduction}
\label{sec:intro}

\begin{figure*}[ht]
  \centering
  \includegraphics{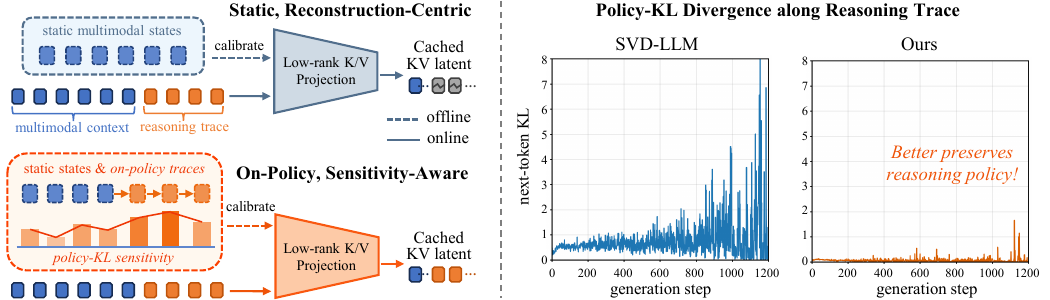}
  \caption{Comparison of compression paradigms and policy-KL divergence during reasoning. \textbf{Left:} Static, reconstruction-centric low-rank KV compression versus our on-policy, sensitivity-aware paradigm. \textbf{Right:} Next-token policy-KL divergence between dense and compressed models along dense-model-generated reasoning traces, evaluated on MMMU dataset using Qwen2.5-VL-7B-Instruct.}
  \label{fig:teaser}
\end{figure*}

Vision-language models (VLMs) are advancing beyond conventional visual understanding and direct answering toward complex multi-step tasks, such as visual mathematical and scientific reasoning~\cite{lu2024mathvista,lu2022scienceqa,yue2024mmmu}. Reasoning VLMs leverage extended chain-of-thought generation to address these challenges~\cite{zhu2025internvl3,bai2025qwen3vl}, but at the cost of substantial inference overhead. In particular, their KV cache must store numerous visual and textual tokens from multimodal input contexts and continuously grow as new reasoning states are generated during autoregressive decoding. The resulting memory storage and bandwidth demands pose significant bottlenecks to efficient long-context inference, necessitating effective KV cache compression techniques.

Existing KV cache compression methods reduce cache size along different axes. Token eviction selectively retains past tokens~\cite{li2024snapkv,cai2026rkv}, while quantization lowers numerical precision to reduce per-token storage cost, but risks substantial accuracy loss at low bit-widths~\cite{liu2024kivi,hooper2024kvquant}. Recently, low-rank compression has emerged as a promising paradigm, factorizing KV projection matrices to cache low-dimensional latents instead of full KV states~\cite{wang2025svdllm,chang2025palu, bhatnagar2026starkv}. Moreover, suitable transformations of these latents can further improve their quantization fidelity, enabling a more favorable accuracy–memory trade-off~\cite{chang2025palu}. 
Existing VLM low-rank methods typically derive low-rank projections by minimizing KV reconstruction error over static multimodal calibration sequences, with evaluations primarily conducted on short-form or direct-answer tasks~\cite{wang2026qsvd,wang2026wsvd}. 

Although previous low-rank methods effectively compress multimodal input states, we notice substantial degradation when applying them to reasoning VLMs, where lengthy generated states must also be compressed. As shown in \cref{fig:teaser} (right), when evaluated on dense-model-generated prefixes, the next-token policy-KL divergence between the dense and compressed models grows sharply as generation length increases. End-to-end generation also reveals reasoning failures, such as logical errors and infinite loops. These findings suggest that static, reconstruction-centric compression does not necessarily preserve the model's predictive policy throughout reasoning, raising a central question: \textbf{How can low-rank KV compression preserve model's reasoning policy while compressing KV states of both large multimodal contexts and growing reasoning traces under a constrained cache budget?}

To understand this degradation, we conduct empirical analyses and identify two key issues: 
\textbf{(1) Calibration-state mismatch:} Static calibration on input multimodal contexts may not adequately cover self-generated reasoning KV states. We observe substantially higher reconstruction errors during reasoning than on input contexts, while incorporating ground-truth solution traces only partially alleviates this mismatch. \textbf{(2) Unequal error sensitivities:} Equal-magnitude errors at different KV states can have substantially different impacts on the model's policy. In particular, many visual-token states exhibit lower policy-KL sensitivity than certain textual and reasoning states, with considerable variation even among reasoning tokens. Consequently, uniform reconstruction may fail to allocate limited low-rank capacity to the most policy-sensitive states.

Motivated by these observations, we propose \textbf{TraceKV}, a training-free framework \textbf{guided by a unified policy-preserving principle} that minimizes policy-KL. \textbf{Based on Observation 1,} we collect calibration activations from both multimodal contexts and on-policy traces generated by the dense VLM, improving alignment with inference-time states. \textbf{Based on Observation 2,} we estimate KV state sensitivities using the local Fisher information of policy-KL and formulate an on-policy, sensitivity-aware surrogate objective solvable by SVD. We further estimate the policy-KL importance of resulting singular components to globally allocate ranks across KV projections and layers. Together, our method addresses both state alignment and sensitivity-aware protection, as illustrated in Fig.~\ref{fig:teaser} (left), substantially reducing policy-KL divergence during reasoning (right).


Our contributions are summarized as follows:
\begin{itemize}
    \item \textbf{Analysis.} We analyze low-rank KV compression on reasoning VLMs and identify two sources of policy drift: calibration-state mismatch and unequal error sensitivity.
    \item \textbf{Method.} We propose TraceKV, a training-free low-rank KV compression framework that unifies on-policy calibration, sensitivity-aware SVD, and global rank allocation under a single policy-preserving principle.
    \item \textbf{Results.} Extensive experiments across multiple reasoning VLMs and visual reasoning benchmarks demonstrate superior accuracy over existing compression baselines at comparable memory budgets. TraceKV achieves 5.64$\times$ KV-cache compression with near-lossless accuracy at 70\% rank ratio with quantization, and up to 1.83$\times$ end-to-end decoding speedup with optimized kernels.
\end{itemize}
\section{Preliminaries}
\label{sec:prelim}

\paragraph{Low-Rank KV Cache Compression.}
Low-rank KV compression alleviates substantial KV cache memory overhead by factorizing the key and value projection matrices and caching low-dimensional latents. For a KV projection matrix $W \in \mathbb{R}^{d_{\mathrm{in}} \times d_{\mathrm{out}}}$, we construct a low-rank approximation $\hat W = AB$, where $A \in \mathbb{R}^{d_{\mathrm{in}} \times r}$ and $B \in \mathbb{R}^{r \times d_{\mathrm{out}}}$, with $r < d_{\mathrm{out}}$. Given a token activation $x_t$, only the low-dimensional latent $z_t=x_tA$ is cached, rather than the full KV state $y_t=x_tW$. This reduces the KV cache memory footprint by storing $r$ instead of $d_{\mathrm{out}}$ elements per token. During attention computation, the pre-RoPE KV state is approximately reconstructed as $\hat y_t = z_tB=x_t\hat W$.

\paragraph{Reconstruction-Based SVD.}
Existing SVD-based compression methods~\cite{wang2025svdllm,chang2025palu} often formulate low-rank approximation as minimizing the \textbf{KV reconstruction error} over calibration activations. Let $X \in \mathbb{R}^{N \times d_{\mathrm{in}}}$ denote the activation matrix collected from a calibration set, where $N$ is the total calibration tokens. The reconstruction objective is

\begin{equation}
\label{eq:recons}
    \min_{\operatorname{rank}(\hat W)\le r}
    \|X(W-\hat W)\|_F^2.
\end{equation}

This objective admits an optimal rank-$r$ solution via whitened SVD~\cite{wang2025svdllm}. Specifically, a whitening transform $S$ is obtained from the Cholesky decomposition of the activation covariance matrix $X^\top X$, followed by SVD of $SW=U\Sigma V^\top$. The optimal rank-$r$ approximation is then given by retaining the first $r$ singular components: $A=S^{-1}U_r\Sigma_r^{1/2}$ and $B=\Sigma_r^{1/2}V_r^\top$, 
yielding $\hat W=AB$. 

\paragraph{Policy-KL Divergence.}
We consider the discrepancy between the original model policy $p$ and the compressed model policy $p_\delta$ over autoregressive reasoning states. Given a fixed prefix $c$, the policy-KL divergence is measured by $D_{\mathrm{KL}}(p(\cdot\mid c)\,\|\,p_\delta(\cdot\mid c))$, where $\delta$ denotes a perturbation to an internal model variable $\theta$. Since the policy-KL divergence vanishes at the original model ($\delta=0$) and its first-order derivative is zero, a second-order Taylor expansion around this point yields
\begin{equation}
\label{eq:kl_second_order}
    D_{\mathrm{KL}}(p\,\|\,p_\delta)
    = \frac{1}{2}\delta^\top F_\theta\delta
    + o(\|\delta\|^2),
\end{equation}
where $F_\theta$ is the Fisher information matrix with respect to $\theta$:
\begin{equation}
\label{eq:fisher}
    F_\theta = \mathbb{E}_{a\sim p(\cdot\mid c)}
    \left[
        \nabla_\theta \log p(a\mid c)\,
        \nabla_\theta \log p(a\mid c)^\top
    \right].
\end{equation}

This establishes Fisher information as a local measure of policy-KL increase to perturbations, providing a principled basis for evaluating the impact of compression.
\section{Motivation}
\label{sec:motiv}

As illustrated in Fig.~\ref{fig:teaser}, we observe a progressively increasing policy-KL divergence between low-rank-compressed and dense models during reasoning. We investigate two underlying causes of this degradation through controlled analyses on Qwen2.5-VL-7B-Instruct~\cite{bai2025qwen25vl}.

\paragraph{Observation 1: Static calibration yields larger errors on reasoning KV states.}
Existing VLM low-rank compression methods typically construct calibration activation matrices from static input multimodal contexts to learn low-rank projections for accurate reconstruction. While effective for short-answer tasks, reasoning VLMs generate lengthy reasoning KV states throughout decoding, which are compressed using the same low-rank projections and repeatedly accessed in subsequent steps. This creates a potential mismatch between the calibrated low-rank subspaces and the actual inference-time reasoning states. 

\begin{figure}[t]
  \centering
  \includegraphics[]{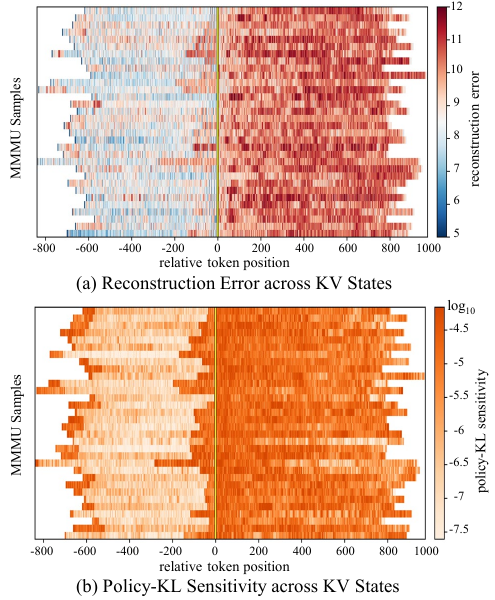}
  \caption{Analysis of KV states on 32 MMMU samples (rows). (a) KV reconstruction errors at 50\% rank ratio under input-context calibration, and (b) KV state sensitivities estimated using Fisher information, averaged over all KV projections. Token positions are aligned at the input-generation boundary (yellow line).}
  \label{fig:motiv}
\end{figure}

To examine this mismatch, \cref{fig:motiv}(a) visualizes the KV reconstruction error, $e_t=\|x_t(W-\hat W)\|_2^2$, averaged over KV projections across all layers at each token position. Token positions are aligned at the input-generation boundary (yellow line), separating the input context tokens and generated tokens. As shown, while errors remain relatively low on input KV states, they increase sharply upon entering the reasoning phase and remain elevated throughout generation. As a result, these errors may accumulate during reasoning and lead to policy drift from the dense model. 

We further compare calibration strategies in \cref{tab:recons_error}. Adding ground-truth solution traces to the calibration activations reduces reasoning-state error from 10.27 to 9.67. However, such static traces remain insufficient to capture the model's actual reasoning distribution. Instead, we augment calibration with activations from self-generated reasoning traces of the dense model (i.e., \textbf{on-policy traces}), further reducing the error to 8.88 with negligible changes in input-context error. These findings motivate on-policy calibration to better preserve reasoning KV states.

\begin{table}[t]
\centering
\caption{Average reconstruction error on MMMU at 50\% rank ratio under different strategies. Calibration data is from ScienceQA. }
\label{tab:recons_error}
\small
\begin{adjustbox}{max width=\linewidth}
\setlength{\tabcolsep}{5pt}
\begin{tabular}{@{}lccc@{}}
\toprule
\textbf{Calibration Data} &
\textbf{Input Context} &
\textbf{Reasoning Trace} &
\textbf{Overall} \\
\midrule
Input Context
& 8.02 & 10.27 & 9.21 \\
+ GT Solutions
& 8.01 & 9.67 & 8.89 \\
+ On-policy Traces
& \textbf{7.96} & \textbf{8.88} & \textbf{8.45} \\
\bottomrule
\end{tabular}
\end{adjustbox}
\end{table}

\begin{figure*}[ht]
  \centering
  \includegraphics{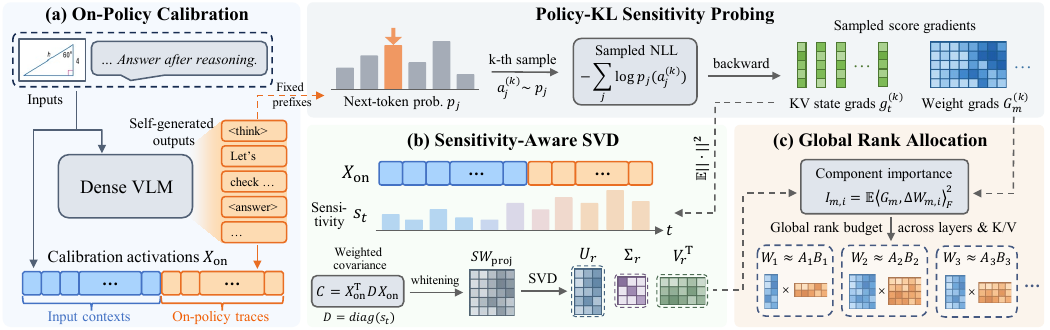}
  \caption{Overview of TraceKV. \textbf{Left:} The dense VLM generates on-policy traces to construct calibration data covering both multimodal contexts and reasoning states. \textbf{Top:} Monte Carlo sampling of next-token predictions provides score gradients for estimating policy-KL sensitivities. \textbf{Bottom:} KV-state sensitivities guide weighted SVD to obtain singular components, which are further evaluated by their policy-KL importance and selected under a shared global rank budget to construct the final low-rank projections.}
  \label{fig:overview}
\end{figure*}

\paragraph{Observation 2: KV states have unequal error sensitivities.}
While incorporating on-policy reasoning traces mitigates the calibration distribution mismatch, the conventional reconstruction objective in \cref{eq:recons} still weights errors across all KV states uniformly. 
However, equal-magnitude KV perturbations may have vastly different impacts on subsequent reasoning: errors in redundant visual states may be inconsequential, whereas those in reasoning-critical states can substantially alter the generation trajectory.

To investigate this heterogeneity, we quantify KV state sensitivities using estimated Fisher information (formally defined in Sec.~\ref{sec:sensitivity}), which approximates the impact of local KV perturbations on the subsequent cumulative policy-KL divergence. As shown in Fig.~\ref{fig:motiv}(b), sensitivity varies substantially across token positions: many visual input states, corresponding to the large central region on the left, exhibit low sensitivity, while certain textual input states and generated reasoning states on the right are considerably more sensitive. Moreover, sensitivity also varies notably within reasoning states.
These findings suggest that the uniform reconstruction objective overlooks the unequal sensitivities of KV states, motivating sensitivity-aware protection.

\section{Method}
\label{sec:method}

The above analyses suggest that the conventional KV reconstruction objective may be suboptimal. Motivated by knowledge distillation~\cite{hinton2015distilling}, we instead seek to preserve the dense model's predictive policy under low-rank compression. For a KV projection matrix $W$ with rank budget $r$, we formulate the \textbf{ideal policy-KL objective} as
\begin{equation}
\label{eq:ideal_kl}
\min_{\operatorname{rank}(\hat W)\le r}
\mathbb{E}_{s\sim d_p}
\left[
D_{\mathrm{KL}}\big(
p(\cdot\mid s)\,\|\,p_{\hat W}(\cdot\mid s)
\big)
\right],
\end{equation}
where $d_p$ denotes the distribution of calibration prefixes along the dense model's on-policy reasoning traces, and $p_{\hat W}$ is the policy induced by replacing $W$ with its low-rank approximation $\hat W$. Since this ideal objective admits no closed-form solution, we instead develop an \textbf{on-policy, sensitivity-aware objective as a tractable surrogate} solvable via SVD. We then extend the policy-preserving principle to global rank allocation across KV projections and layers. The overall framework is illustrated in \cref{fig:overview}.

\subsection{On-Policy Calibration}

Observation 1 reveals that calibration with on-policy reasoning traces yields lower reconstruction errors than using static input contexts, even when augmented with ground-truth solution traces. This suggests a distribution mismatch between predefined calibration data and the states encountered during the model's own reasoning process.

Motivated by this finding, we let the dense VLM autoregressively generate reasoning traces for image-question pairs in the calibration set. We then collect activations from both the multimodal input contexts and the self-generated reasoning states, forming the \textbf{on-policy calibration matrix $X_{\text{on}}$} for each KV projection. Replacing the static calibration matrix $X$ in \cref{eq:recons} with $X_{\text{on}}$ yields 
\begin{equation}
\label{eq:recons_1}
\min_{\operatorname{rank}(\hat W)\le r}
||X_{\text{on}}(W-\hat W)||_F^2.
\end{equation}

The generated prefixes are also retained for subsequent policy-KL sensitivity estimation. This procedure aligns calibration with the policy-KL objective in \cref{eq:ideal_kl}, as the next-token policy depends on both the multimodal input and the evolving reasoning prefix, where the collected on-policy prefixes provide an empirical approximation to $d_p$.

\subsection{Sensitivity-aware SVD}
\label{sec:sensitivity}
Observation 2 reveals substantial heterogeneity in how KV-state perturbations affect the reasoning policy. We first formalize the sensitivity measure used in the analysis. 

A perturbation to a KV state can affect predictions at all subsequent positions. Building on the second-order policy-KL approximation in \cref{eq:kl_second_order,eq:fisher}, let $p_j=p(\cdot\mid c_j)$ denote the dense policy at prefix $c_j$, and $h_{t,j}(a)=\nabla_{y_t}\log p_j(a)$ denote the score gradient with respect to a KV state $y_t$. We define its policy-KL sensitivity $s_t$ as

\begin{equation}
\label{eq:sensitivity_0}
\begin{aligned}
F_t = \sum_{j\in\mathcal J(t)}
\mathbb E_{a\sim p_j}
[h_{t,j}(a)h_{t,j}(a)^\top],\quad
s_t = \frac{1}{2d_{\mathrm{out}}}\operatorname{Tr}(F_t),
\end{aligned}
\end{equation}
where $\mathcal J(t)$ denotes all prediction positions after $y_t$. Intuitively, since $s_t=\frac{1}{2d_{\mathrm{out}}}\operatorname{Tr}(F_t)=\mathbb E_{||u||=1}[\frac{1}{2}u^\top F_tu]$, according to \cref{eq:kl_second_order}, $s_t$ quantifies the \textbf{expected second-order increase in cumulative policy-KL under an isotropic, unit-energy perturbation $u$ to $y_t$.}

Since explicitly computing the Fisher matrices $F_t$ is prohibitively expensive, we employ Monte Carlo estimation. For each fixed on-policy prefix $c_j$, we independently sample $a_j^{(k)}\sim p_j$ and construct NLL $ \mathcal L^{(k)}=-\sum_j\log p_j(a_j^{(k)}) $. 
The resulting gradients $g_t^{(k)}=\nabla_{y_t}\mathcal L^{(k)}$ yield
\begin{equation}
\label{eq:sensitivity}
\operatorname{Tr}(F_t)= \mathbb E||g_t||_2^2,\quad
\hat s_t
=\frac{1}{2Kd_{\mathrm{out}}}
\sum_{k=1}^{K}||g_t^{(k)}||_2^2,
\end{equation}
where $K$ is the number of Monte Carlo samples. This estimator is unbiased since independently sampled score gradients across prediction positions have zero mean, causing cross-position terms to vanish in expectation. 

The pronounced sensitivity variation revealed by Observation 2 motivates prioritizing fidelity for reasoning-critical KV states. Accordingly, we extend \cref{eq:recons_1} to the final \textbf{on-policy, sensitivity-aware objective}:
\begin{equation}
\label{eq:recons_2}
\min_{\operatorname{rank}(\hat W)\le r}
||D^{1/2}X_{\text{on}}(W-\hat W)||_F^2,
\end{equation}
where $D=\operatorname{diag}(\hat s_1,\ldots,\hat s_N)/(\sum \hat s)$ weights error of each KV state by its normalized policy-KL sensitivity. Similar to reconstruction-based SVD in \cref{sec:prelim}, this objective admits an optimal rank-$r$ solution. We first construct the weighted covariance $C=X_{\text{on}}^\top DX_{\text{on}}$ and obtain its Cholesky factor $S^\top S=C$. We then perform SVD on the whitened weight matrix $SW=U\Sigma V^\top$, yielding a set of singular components $\Delta W_i=S^{-1}u_i\sigma_i v_i^\top$. The leading $r$ components provide an optimal rank-$r$ solution to the above objective.

We further show \cref{eq:recons_2} as a \textbf{tractable surrogate of the ideal policy-KL objective} in \cref{eq:ideal_kl} and analyze the associated approximation errors in \cref{app:surrogate}.

\subsection{Global Rank Allocation}

The preceding method yields a fixed-rank surrogate solution for each KV projection by retaining its leading singular components. However, prior studies have shown that using uniform ranks across layers and KV projections under a global budget can be suboptimal, as their impact on model performance varies substantially~\cite{chang2025palu,abbasi2026zssvd}. 

Following the same policy-preserving principle, we again leverage the Fisher quadratic approximation in \cref{eq:kl_second_order,eq:fisher} to estimate the \textbf{second-order policy-KL increase of removing a singular component} as its importance. Let $m$ index KV projections across all layers, and $i$ index the singular components $\Delta W_{m,i}$ of projection $W_m$. Reusing the sampled NLL $\mathcal L^{(k)}$ in \cref{sec:sensitivity}, we compute $G_m^{(k)}=\nabla_{W_m}\mathcal L^{(k)}$ and estimate the policy-KL importance of each singular component as

\begin{equation}
\label{eq:importance}
I_{m,i}=\frac{1}{2K}\sum_{k=1}^{K}
\left\langle G_m^{(k)},\Delta W_{m,i}\right\rangle_F^2.
\end{equation}

We then globally select singular components with the highest importance scores under the total rank budget. Let $\mathcal I_m$ denote the selected component indices for projection $W_m$, determining its rank $r_m=|\mathcal I_m|$. The final low-rank factors are $A_m=S_m^{-1}U_{\mathcal I_m}\Sigma_{\mathcal I_m}^{1/2}$ and $B_m=\Sigma_{\mathcal I_m}^{1/2}V_{\mathcal I_m}^{\top}$.

\subsection{Low-Bit Quantization Integration}
To further reduce KV cache memory, we apply low-bit quantization to the low-rank latents. The uneven singular values from SVD can concentrate energy in leading latent dimensions, introducing outliers that hinder quantization. Following prior works~\cite{chang2025palu,wang2026wsvd}, we apply an orthogonal Hadamard transform to balance latent magnitudes, similar to QuaRot~\cite{ashkboos2024quarot}. This transformation can be fused into the low-rank factors without additional inference overhead. Further quantization details are provided in \cref{app:quantization}.



\subsection{System Implementation}
\label{sec:system}

Naively reconstructing and materializing the full KV cache at every decoding step would incur substantial computation and memory traffic. Thanks to our regular latent-cache structure and fixed computation pipeline, we develop an efficient low-rank attention implementation that translate memory savings into practical decoding acceleration. 

For values, we aggregate low-rank latents using attention weights before reconstruction, avoiding full-sequence value-cache reconstruction. For keys, reconstruction must precede applying RoPE. We therefore implement custom Triton kernels that fuse key reconstruction, RoPE, and attention-logit computation, avoiding intermediate key-cache materialization and reducing memory traffic in bandwidth-bound decoding. For 4-bit quantization, dequantization is further fused into these kernels. Additional implementation details are provided in \cref{app:attention_impl}
.    
\begin{table*}[ht]
\centering
\caption{Accuracy (\%) comparison of low-rank only KV compression methods on multimodal reasoning benchmarks. The rank ratio denotes the fraction of retained KV dimensions across all heads and layers.}
\label{tab:low_rank_main}
\small
\setlength{\tabcolsep}{4pt}
\begin{adjustbox}{max width=\textwidth}
\begin{tabular}{llcccccccccc}
\toprule
\multirow{2}{*}{\textbf{Method}} &
\multirow{2}{*}{\textbf{Rank}} &
\multicolumn{3}{c}{\textbf{Qwen2.5-VL-7B}} &
\multicolumn{3}{c}{\textbf{InternVL3-8B}} &
\multicolumn{3}{c}{\textbf{Qwen3-VL-8B}} &
\multirow{2}{*}{\textbf{Overall Avg.}} \\
\cmidrule(lr){3-5}
\cmidrule(lr){6-8}
\cmidrule(lr){9-11}
& & SciQA & MathV & MMMU
  & SciQA & MathV & MMMU
  & SciQA & MathV & MMMU & \\
\midrule
Dense & 100\%
& 87.85 & 67.3 & 53.00
& 93.16 & 70.1 & 58.56
& 93.60 & 77.0 & 61.67 & 73.58 \\
\midrule

SVD-LLM & 70\%
& 87.56 & 62.4 & 46.67
& 93.45 & 65.7 & 54.88
& 89.49 & 71.0 & 54.11 & 69.47 \\
ZS-SVD & 70\%
& 85.18 & 58.0 & 45.44
& \textbf{93.90} & 68.5 & 57.67
& 91.13 & 72.1 & 54.33 & 69.58 \\
Palu & 70\%
& 83.54 & 63.8 & 50.89
& 93.69 & 67.6 & 58.22
& 90.98 & 74.1 & 59.11 & 71.33 \\
WSVD & 70\%
& 85.18 & 61.0 & 47.22
& 91.27 & 63.8 & 56.33
& 93.46 & 74.7 & 58.11 & 70.12 \\
\textbf{Ours} & 70\%
& \textbf{87.85} & \textbf{67.7} & \textbf{51.33}
& 93.60 & \textbf{69.9} & \textbf{59.11}
& \textbf{93.95} & \textbf{76.1} & \textbf{60.44} & \textbf{73.33} \\
\midrule

SVD-LLM & 50\%
& 82.75 & 54.6 & 37.89
& 86.86 & 51.1 & 44.66
& 79.38 & 59.4 & 40.22 & 59.65 \\
ZS-SVD & 50\%
& 83.09 & 49.9 & 42.56
& 89.09 & 58.5 & 51.33
& 66.24 & 41.9 & 21.89 & 56.06 \\
Palu & 50\%
& 74.17 & 55.1 & 43.77
& 92.61 & 59.9 & 52.89
& 83.39 & 61.3 & 45.56 & 63.19 \\
WSVD & 50\%
& 71.64 & 40.0 & 37.33
& 81.41 & 36.1 & 36.44
& 84.78 & 51.5 & 38.89 & 53.12 \\
\textbf{Ours} & 50\%
& \textbf{87.40} & \textbf{63.2} & \textbf{49.11}
& \textbf{92.96} & \textbf{65.0} & \textbf{54.67}
& \textbf{92.76} & \textbf{72.6} & \textbf{54.78} & \textbf{70.28} \\
\bottomrule
\end{tabular}
\end{adjustbox}
\end{table*}
\begin{table*}[t!]
\centering
\caption{
Accuracy (\%) comparison of KV cache compression methods with quantization.
Comp. denotes the KV cache compression factor relative to the original BF16 cache (including quantization metadata).
Overall Avg. is averaged across all models and benchmarks.}
\label{tab:quant_main}
\small
\setlength{\tabcolsep}{4pt}
\begin{adjustbox}{max width=\textwidth}
\begin{tabular}{lcccccccccccc}
\toprule
\multirow{2}{*}{\textbf{Method}} &
\multirow{2}{*}{\textbf{Bits}} &
\multirow{2}{*}{\textbf{Comp.}} &
\multicolumn{3}{c}{\textbf{Qwen2.5-VL-7B}} &
\multicolumn{3}{c}{\textbf{InternVL3-8B}} &
\multicolumn{3}{c}{\textbf{Qwen3-VL-8B}} &
\multirow{2}{*}{\textbf{Overall Avg.}} \\
\cmidrule(lr){4-6}
\cmidrule(lr){7-9}
\cmidrule(lr){10-12}
& & & SciQA & MathV & MMMU
  & SciQA & MathV & MMMU
  & SciQA & MathV & MMMU & \\
\midrule
Dense & 16 & 1.00$\times$ 
& 87.85 & 67.3 & 53.00 & 93.16 & 70.1 & 58.56 & 93.60 & 77.0 & 61.67 & 73.58 \\
\midrule
RTN-KV4 (gs32) & 4 & 3.20$\times$
& 46.60 & 47.8 & 28.44 & 77.24 & 44.9 & 46.33 & 93.16 & \textbf{76.9} & 58.78 & 57.79 \\
KIVI-2 (gs32, r32) & 2 & 5.33$\times$
& 86.96 & 65.8 & 51.11 & \textbf{94.05} & 69.0 & 56.33 & 91.77 & 76.0 & 58.56 & 72.18 \\
\midrule
Palu (70\%, KV4) & 4 & 5.64$\times$
& 86.07 & 61.5 & 48.56 & 93.15 & 67.0 & 57.33 & 91.27 & 73.8 & 56.67 & 70.59 \\
\textbf{Ours (70\%, KV4)} & 4 & 5.64$\times$
& 88.10 & \textbf{67.5} & \textbf{51.78} & 93.55 & \textbf{69.6} & \textbf{58.67} & 93.65 & 76.7 & \textbf{61.11} & \textbf{73.41} \\
\textbf{Ours (70\%, KV3)} & 3 & 7.44$\times$
& \textbf{88.20} & 65.5 & 49.67 & 93.60 & 68.1 & 58.33 & \textbf{93.80} & 74.7 & 58.22 & 72.24 \\
\midrule
Palu (50\%, KV4) & 4 & 7.76$\times$
& 76.95 & 52.9 & 41.11 & 92.40 & 61.1 & 52.11 & 83.49 & 61.6 & 45.33 & 63.00 \\
\textbf{Ours (50\%, KV4)} & 4 & 7.76$\times$
& 87.41 & 62.7 & 49.89 & 93.46 & 64.8 & 53.56 & 92.41 & 72.0 & 53.11 & 69.93 \\
\textbf{Ours (50\%, KV3)} & 3 & 10.24$\times$
& 87.36 & 62.5 & 47.56 & 92.76 & 64.3 & 52.33 & 91.67 & 67.5 & 50.44 & 68.49 \\
\bottomrule
\end{tabular}
\end{adjustbox}
\end{table*}

\section{Experiments}
\label{sec:experiments}

\subsection{Experimental Setup}

\paragraph{Models and Benchmarks.}
We evaluate our method on 4 reasoning-capable VLMs: Qwen2.5-VL-7B-Instruct~\cite{bai2025qwen25vl}, InternVL3-8B~\cite{zhu2025internvl3}, and Qwen3-VL-8B/32B-Thinking~\cite{bai2025qwen3vl}.
We evaluate on three multimodal reasoning benchmarks: ScienceQA~\cite{lu2022scienceqa} for scientific reasoning, MathVista~\cite{lu2024mathvista} for mathematical reasoning, and MMMU~\cite{yue2024mmmu} for multidisciplinary reasoning over visual contexts.
All evaluations are conducted using the lmms-eval framework~\cite{zhang2025lmmseval} with reasoning-oriented prompts.
Further details of the reasoning evaluation protocols are provided in \cref{app:protocols}.

\paragraph{Baselines.}
We compare against general-purpose SVD-based compression methods, including SVD-LLM~\cite{wang2025svdllm} and ZS-SVD~\cite{abbasi2026zssvd}, as well as low-rank KV cache compression methods, including Palu~\cite{chang2025palu} and WSVD~\cite{wang2026wsvd}. For a fair comparison, we restrict all methods to compressing only the key and value projection matrices.
We also compare against two KV cache quantization baselines: token-wise RTN quantization and KIVI~\cite{liu2024kivi}. Token eviction methods are excluded as they are largely orthogonal.

\paragraph{Implementation Details.}
For baselines, we follow the official implementation of WSVD~\cite{wang2026wsvd}, which randomly samples 256 examples from the ScienceQA train split, collects calibration activations on input image-text contexts, and estimates calibration loss or Fisher information using ground-truth solutions. Our method uses the same calibration samples with the proposed on-policy, sensitivity-aware calibration.
Following Palu~\cite{chang2025palu}, we adopt a head-group size of 4 for low-rank decomposition. 
When combined with quantization, our method employs simple per-token-latent asymmetric quantization without further grouping.
All experiments are conducted on NVIDIA L40S GPUs.

\subsection{Main Results}

\paragraph{Low-Rank KV Compression.}
Table~\ref{tab:low_rank_main} compares low-rank-only KV compression methods.
At 70\% rank ratio, our method achieves nearly lossless compression, retaining 73.33\% overall accuracy compared to 73.58\% for the dense models.
At 50\% rank ratio, our method maintains 70.28\% accuracy, surpassing the best baseline, Palu, by \textbf{7.09} p.p.
The larger degradation of SVD-LLM and Palu suggests that static calibration and reconstruction-centric objectives are insufficient to preserve reasoning performance.
WSVD suffers even more severe degradation, potentially due to its per-head SVD limiting cross-head component sharing~\cite{chang2025palu}.
Additional results under alternative baseline calibration protocols are provided in \cref{app:baseline_calibration}.

\paragraph{Integration with Quantization.}
\cref{tab:quant_main} compares KV quantization and low-rank KV compression combined with quantization. 
Naive token-wise RTN quantization at 4-bit leads to substantial accuracy degradation on Qwen2.5-VL-7B and InternVL3-8B.
In contrast, our low-rank projections combined with Hadamard transform produce quantization-friendly KV latents, where 4-bit quantization introduces negligible additional accuracy loss over low-rank compression alone. At 70\% rank ratio with 4-bit KV quantization, our method achieves 73.41\% overall accuracy compared to 73.58\% for the dense model, while reducing KV cache size by \textbf{5.64$\times$}.
This demonstrates the complementary benefits of low-rank compression and quantization.
Even under aggressive 3-bit quantization, our method maintains competitive reasoning performance.
Compared to KIVI, we achieve higher accuracy at comparable KV memory budgets without requiring per-channel key quantization or fine-grained element grouping.

\begin{table}[t]
\centering
\caption{Results on Qwen3-VL-32B-Thinking.
Comp. denotes the KV cache compression factor.}
\label{tab:qwen3_32b}
\small
\begin{adjustbox}{max width=\linewidth}
\setlength{\tabcolsep}{3pt}
\begin{tabular}{@{}lccccc@{}}
\toprule
\textbf{Method} & \textbf{Comp.} & \textbf{SciQA} &
\textbf{MathV} & \textbf{MMMU} & \textbf{Avg.} \\
\midrule
Dense                  & 1.00$\times$ & 93.80 & 80.4 & 66.67 & 80.29 \\
\midrule
SVD-LLM (50\%)         & 2.00$\times$ & 93.95 & 70.8 & 59.44 & 74.73 \\
Palu (50\%)            & 2.00$\times$ & 91.27 & 75.9 & 55.56 & 74.24 \\
\textbf{Ours (50\%)}   & 2.00$\times$ & 94.00 & 78.6 & 64.00 & 78.87 \\
\midrule
Palu (50\%, KV4)       & 7.76$\times$ & 90.23 & 75.5 & 56.11 & 73.95 \\
\textbf{Ours (50\%, KV4)}
                       & 7.76$\times$ & 93.95 & 79.2 & 63.78 & 78.98 \\
\bottomrule
\end{tabular}
\end{adjustbox}
\end{table}
\begin{table}[t]
\centering
\caption{Component ablation on Qwen2.5-VL-7B-Instruct
at 50\% rank ratio without KV quantization.}
\label{tab:ablation}
\small
\begin{adjustbox}{max width=\linewidth}
\setlength{\tabcolsep}{3pt}
\begin{tabular}{@{}lcccc@{}}
\toprule
\textbf{Variant} &
\textbf{SciQA} &
\textbf{MathV} &
\textbf{MMMU} &
\textbf{Avg.} \\
\midrule
Baseline (SVD-LLM)
& 82.75 & 54.60 & 37.89 & 58.41 \\
+ GT Solutions
& 83.74 & 57.20 & 40.89 & 60.61 \\
+ On-policy Traces
& 85.57 & 57.60 & 43.11 & 62.09 \\
+ On-policy \& Sensitivity
& 86.81 & 58.80 & 45.44 & 63.68 \\
\textbf{Ours (Full)}
& \textbf{87.40} & \textbf{63.20}
& \textbf{49.11} & \textbf{66.57} \\
\bottomrule
\end{tabular}
\end{adjustbox}
\end{table}

\paragraph{Evaluation on Larger Models.}
We further evaluate representative methods on Qwen3-VL-32B-Thinking~\cite{bai2025qwen3vl}, as shown in \cref{tab:qwen3_32b}. Notably, with 50\% rank ratio and 4-bit latent quantization, TraceKV achieves an average accuracy of 78.98\%, only 1.31 percentage points below the dense model. Interestingly, the accuracy degradation is smaller than that observed on Qwen3-VL-8B-Thinking under the same configuration, suggesting that larger reasoning VLMs may tolerate more aggressive KV compression. These results highlight our method's scalability for efficient inference at larger model scales.

\subsection{Ablation and Analysis}

\paragraph{Component Ablation.}
\cref{tab:ablation} evaluates the contribution of each component of our method.
To isolate the effect of rank allocation, we use uniform ranks across projections for all variants except the full method.
Incorporating ground-truth solution traces into calibration statistics only moderately improves average accuracy from 58.41\% to 60.61\%.
In contrast, calibrating on on-policy reasoning traces achieves a larger improvement to 62.09\%, highlighting the importance of capturing the actual inference-time reasoning distribution.
Sensitivity-aware SVD further increases accuracy to 63.68\% by prioritizing reasoning-critical KV states during compression.
Finally, our global rank allocation brings the full method to 66.57\% accuracy, outperforming the baseline by \textbf{8.16} p.p.

\paragraph{Compression Error Analysis.}

\begin{figure}[t]
    \centering
    \includegraphics[]{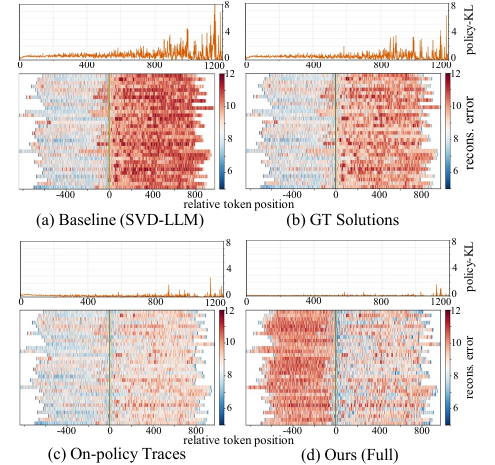}
    \caption{Visualization of policy-KL divergence and reconstruction error under different compression strategies. Evaluated using Qwen2.5-VL-7B-Instruct at 50\% rank ratio on MMMU.}
    \label{fig:error_vis}
\end{figure}

To intuitively understand the advantages of our proposed methods, \cref{fig:error_vis} visualizes the policy-KL divergence and token-wise reconstruction error under different compression strategies.
As shown in (a), the SVD-LLM baseline exhibits low reconstruction error on the input context but a sharp increase upon entering the reasoning phase, accompanied by rapidly growing policy-KL throughout generation.
Incorporating ground-truth solution traces in (b) partially alleviates these issues.
By calibrating on on-policy reasoning traces, (c) substantially reduces both policy-KL divergence and reconstruction error  during later generation steps.
Finally, incorporating sensitivity-aware SVD and global rank allocation in (d) redistributes fidelity: errors increase on less sensitive visual states, while critical KV states are preserved with lower errors (deeper blue). Notably, we observe that initial sink tokens consistently exhibit very low errors due to their high sensitivity, consistent with prior observations and methods that explicitly preserve sink tokens~\cite{hooper2024kvquant, xiao2024streamingllm}.
Consequently, our full method exhibits the slowest growth in policy-KL divergence, better preserving the dense model's reasoning policy.

\paragraph{Calibration Domain Shift.}
To assess robustness to calibration domain shift, we further evaluate the use of an alternative calibration dataset.
We sample 256 examples from a held-out split of MathVista and follow the same calibration procedure, with results reported in \cref{tab:calib_set}.
As shown, in-domain calibration yields higher accuracy compared to cross-domain calibration, improving our method by 2.33 and 2.10 percentage points on ScienceQA and MathVista, respectively. 
Nevertheless, our method consistently outperforms the baseline across all three evaluation benchmarks regardless of the calibration dataset, demonstrating its effectiveness under domain shift.
This cross-domain effectiveness is further supported by the component ablation in \cref{tab:ablation} and the visualization analysis in \cref{fig:error_vis}, which also demonstrate improvements on MathVista and MMMU despite using ScienceQA for calibration.

\begin{table}[t]
\centering
\caption{Accuracy (\%) with different calibration
sets on Qwen2.5-VL-7B-Instruct at 50\% rank ratio.}
\label{tab:calib_set}
\small
\begin{adjustbox}{max width=\linewidth}
\setlength{\tabcolsep}{4pt}
\begin{tabular}{@{}llcccc@{}}
\toprule
\textbf{Calib. Set} &
\textbf{Method} &
\textbf{SciQA} &
\textbf{MathV} &
\textbf{MMMU} &
\textbf{Avg.} \\
\midrule
\multirow{2}{*}{ScienceQA}
& SVD-LLM
& 82.75 & 54.60 & 37.89 & 58.41 \\
& \textbf{Ours}
& \textbf{87.40} & 63.20 & \textbf{49.11} & \textbf{66.57} \\
\midrule
\multirow{2}{*}{MathVista}
& SVD-LLM
& 81.70 & 59.60 & 39.33 & 60.21 \\
& \textbf{Ours}
& 85.07 & \textbf{65.30} & 48.11 & 66.16 \\
\bottomrule
\end{tabular}
\end{adjustbox}
\end{table}

\subsection{System Efficiency}

\begin{figure}[t]
    \centering
    \includegraphics[]{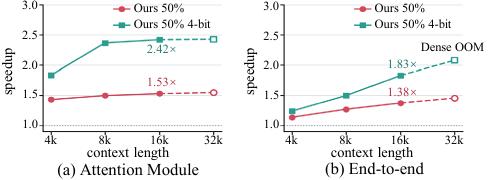}
    \caption{Decoding speedup on Qwen2.5-VL-7B (batch size 16). At 32K, dense model runs out of memory; dashed lines and hollow markers denote speedups estimated by linear extrapolation.}
    \label{fig:efficiency}
\end{figure}

We measure decoding latency on Qwen2.5-VL-7B-Instruct with a batch size of 16 and context lengths ranging from 4K to 32K. The dense model uses the default Hugging Face implementation with FlashAttention-2~\cite{dao2024flashattention}, while our method employs 50\% low-rank KV compression with the optimized kernels described in \cref{sec:system}. \cref{fig:efficiency} presents the attention-module and end-to-end speedups across different context lengths. At 16K context length, our method achieves 1.53$\times$ attention speedup and 1.38$\times$ end-to-end speedup, further increasing to \textbf{2.42$\times$} and \textbf{1.83$\times$} with 4-bit quantization, while reducing peak GPU memory usage from 32.13 GB to 18.57 GB. 
These gains arise from reduced KV cache memory traffic and efficient low-rank attention execution, highlighting the potential of our approach for efficient long-context and long-reasoning inference. 

\section{Related Work}
\label{sec:related}

\paragraph{KV Cache Compression.}
KV cache compression has been explored through architectural designs and post-training techniques. Multi-Head Latent Attention~\cite{liu2024deepseek} (MLA) incorporates learned low-rank KV bottlenecks into model architectures. In contrast, post-training methods reduce KV memory through token eviction~\cite{li2024snapkv,cai2026rkv}, low-bit quantization~\cite{liu2024kivi,hooper2024kvquant}, or low-rank projection~\cite{chang2025palu,wang2026wsvd}. 

\paragraph{Low-Rank Compression Methods.}
SVD-LLM~\cite{wang2025svdllm} improves over previous SVD-based methods~\cite{hsu2022fwsvd,yuan2023asvd} through activation whitening, while subsequent methods further explore adaptive rank allocation~\cite{chang2025palu,wang2026qsvd,abbasi2026zssvd}.
However, their static calibration and reconstruction-centric objectives do not necessarily preserve reasoning policy.
Recent and concurrent methods explore alternative objectives, such as supervised task loss~\cite{wang2026laser} or attention-level output~\cite{liu2026belt}. We instead leverage policy-KL sensitivity to guide low-rank factorization and rank allocation during VLM reasoning.

\paragraph{Calibration for Post-Training Compression.}
While conventional methods typically rely on static calibration data, Self-calibration~\cite{williams2025self} and  RAC~\cite{lucas2026rac} incorporate model-generated data and on-policy traces into calibration for model pruning. 
Separately, FWSVD~\cite{hsu2022fwsvd} and MBQ~\cite{li2025mbq} leverage loss-based sensitivities for weight compression and VLM quantization, respectively.
Our work investigates related challenges in the context of low-rank KV compression for reasoning VLMs, using a unified policy-preserving principle to explain and guide our method.
\section{Conclusion}
\label{sec:conclusion}

We introduce \textbf{TraceKV}, a training-free low-rank KV cache compression framework for reasoning VLMs that employs on-policy, sensitivity-aware compression for preserving model policy. Extensive experiments demonstrate superior accuracy–memory trade-offs, effective integration with quantization, and practical decoding speedup, highlighting its potential for efficient reasoning VLM inference.

{
    \small
    \bibliographystyle{ieeenat_fullname}
    \bibliography{main}
}

\clearpage
\clearpage
\setcounter{page}{1}
\maketitlesupplementary

\appendix
\section{Theoretical Analysis}
\label{app:theory}

\subsection{KV-State Sensitivity}
\label{app:kv_sensitivity}

We revisit the derivation of KV-state sensitivity in \cref{sec:sensitivity} and establish the unbiasedness of Monte Carlo estimation.

Consider a perturbation $\delta {y_t}$ to a KV state $y_t$. Let $p_j=p(\cdot\mid c_j)$ and $p_{j,\delta}$ denote the dense and perturbed policies at prefix $c_j$, respectively. Aggregating the policy-KL divergence over subsequent prediction positions and applying the second-order approximation in \cref{eq:kl_second_order}, we obtain

\begin{equation}
\label{eq:app:kv_fisher}
\begin{aligned}
\mathcal D_t(\delta y_t)
&=\sum_{j\in\mathcal J(t)}
D_{\mathrm{KL}}(p_j\|p_{j,\delta})
=\frac12\delta y_t^\top F_t\delta y_t+o(\|\delta y_t\|^2),\\
F_t&=\sum_{j\in\mathcal J(t)}
\mathbb E_{a_j\sim p_j}
[h_{t,j}(a_j)h_{t,j}(a_j)^\top],
\end{aligned}
\end{equation}
where $\mathcal J(t)$ denotes all prediction positions after $y_t$, $h_{t,j}(a)=\nabla_{y_t}\log p_j(a)$. For an isotropic, unit-energy perturbation $u$ to $y_t$, uniformly sampled from the unit sphere, we have $\mathbb E[uu^\top]=I/d_{\mathrm{out}}$. The expected second-order policy-KL increase is therefore

\begin{equation}
\mathbb E_{\|u\|=1}\left[\frac12 u^\top F_t u\right]
=\frac{1}{2d_{\mathrm{out}}}\operatorname{Tr}(F_t)
=s_t,
\end{equation}
recovering the sensitivity definition in \cref{eq:sensitivity_0}.

To estimate $s_t$, we independently sample $a_j^{(k)}\sim p_j$ at each fixed on-policy prefix and compute the NLL gradient $g_t^{(k)}=-\sum_{j}h_{t,j}(a_j^{(k)})$. Since score gradients have zero expectation, $\mathbb E_{a_j\sim p_j}[h_{t,j}(a_j)]=0$, independence across sampled tokens eliminates the cross-position terms:

\begin{equation}
\begin{aligned}
\mathbb E\|g_t\|_2^2
=\sum_{j\in\mathcal J(t)}
\mathbb E\|h_{t,j}(a_j)\|_2^2
=\operatorname{Tr}(F_t).
\end{aligned}
\end{equation}
Consequently, the estimator in \cref{eq:sensitivity} satisfies

\begin{equation}
\mathbb E[\hat s_t]
=\frac{1}{2Kd_{\mathrm{out}}}
\sum_{k=1}^{K}\mathbb E\|g_t\|_2^2
=s_t,
\end{equation}
proving that the Monte Carlo estimator is unbiased.

\subsection{Singular-Component Importance}
\label{app:component_importance}

We derive the policy-KL importance of singular components used for global rank allocation in \cref{eq:importance}. For a KV projection $W_m$, removing a singular component $\Delta W_{m,i}$ corresponds to a weight perturbation $-\Delta W_{m,i}$. Following the second-order policy-KL approximation, the resulting policy divergence over calibration prefixes is

\begin{equation}
\begin{aligned}
\mathcal D_{m,i}
&=\sum_j D_{\mathrm{KL}}
\big(p_j\|p_{j,-\Delta W_{m,i}}\big)\\
&\approx \frac12
\operatorname{vec}(\Delta W_{m,i})^\top
F_m\operatorname{vec}(\Delta W_{m,i}),
\end{aligned}
\end{equation}

where $p_{j,-\Delta W_{m,i}}$ denotes the perturbed policy and $F_m$ is the Fisher matrix with respect to $W_m$. Let $h_{m,j}(a_j)=\nabla_{W_m}\log p_j(a_j)$. Expanding $F_m$ using \cref{eq:fisher} yields the component importance

\begin{equation}
\label{eq:app:component_importance}
I_{m,i}=\frac12\sum_j
\mathbb E_{a_j\sim p_j}
\left\langle h_{m,j}(a_j),
\Delta W_{m,i}\right\rangle_F^2
.
\end{equation}

As in KV-state sensitivity estimation, we use independently sampled next-token actions and the aggregated NLL gradients $G_m^{(k)}=-\sum_j h_{m,j}(a_j^{(k)})$ to obtain

\begin{equation}
\hat I_{m,i}=\frac{1}{2K}\sum_{k=1}^{K}
\left\langle G_m^{(k)},\Delta W_{m,i}\right\rangle_F^2.
\end{equation}

The estimation is unbiased by the same zero-mean score property and independence argument established in \cref{app:kv_sensitivity}. Thus, $I_{m,i}$ measures the second-order policy-KL impact of removing a singular component. Although jointly removing multiple singular components may introduce cross terms in the second-order policy-KL approximation, we adopt individual importance scores as a tractable additive proxy for global rank allocation.

\subsection{Policy-KL Surrogate and Approx Errors}
\label{app:surrogate}
While the main paper develops our on-policy, sensitivity-aware objective in \cref{eq:recons_2} based on our two observations, in this section, 
we reversely derive how this objective serves as a surrogate of the ideal policy-KL objective in \cref{eq:ideal_kl} for each KV projection matrix, and characterize the associated approximation errors.

Consider a KV projection $W$ and its low-rank approximation $\hat W=W+\Delta\hat W$. Using the collected on-policy calibration prefixes $\{c_j\}_{j=1}^N$ to approximate the prefix distribution $d_p$, we define the empirical policy-KL divergence as $\mathcal D(W')=\sum_j D_{\mathrm{KL}}(p_j\|p_{j,W'})$. Expanding around the dense weights $W$ yields

\begin{equation}
\label{eq:app:kl_expansion}
\begin{aligned}
\mathcal D(W+\Delta\hat W)
={}&\frac12\operatorname{vec}(\Delta\hat W)^\top
F_W\operatorname{vec}(\Delta\hat W)\\
&+o(\|\Delta\hat W\|_F^2)\\
={}&\frac12\sum_{t,u}\delta y_t^\top F_{tu}\delta y_u
+o(\|\Delta\hat W\|_F^2),
\end{aligned}
\end{equation}
where $\delta y_t=(x_t\Delta\hat W)^\top$ is the perturbation to KV state $y_t$, $F_W$ is the Fisher matrix with respect to $W$, and $F_{tu}$ denotes the $(t,u)$ block of the joint Fisher matrix over KV states. The second equality expresses the weight perturbation through its induced KV-state perturbations.

To derive \cref{eq:recons_2} from this form, two approximations are introduced. First, we neglect cross-state interactions by retaining only the diagonal Fisher blocks $F_{tt}$. Second, we approximate each block isotropically as $\frac{1}{2}F_{tt}\approx s_tI$, where $s_t=\operatorname{Tr}(F_{tt})/(2d_{\mathrm{out}})$. Consequently,

\begin{equation}
\label{eq:app:surrogate}
\begin{aligned}
\mathcal D(W+\Delta\hat W)
&\approx\frac12\sum_t\delta y_t^\top F_{tt}\delta y_t\\
&\approx\sum_t s_t\|\delta y_t\|_2^2\\
&=\|D^{1/2}X_{\mathrm{on}}(W-\hat W)\|_F^2,
\end{aligned}
\end{equation}
where $D=\operatorname{diag}(s_1,\ldots,s_N)$. This recovers our on-policy, sensitivity-aware objective in \cref{eq:recons_2} as a surrogate.

More precisely, denoting this surrogate by $\mathcal R(\hat W)$, the approximation error terms are

\begin{equation}
\label{eq:app:approx_error}
\begin{aligned}
&\mathcal D(W+\Delta\hat W)-\mathcal R(\hat W)\\
={}&\underbrace{\frac12\sum_{t\ne u}
\delta y_t^\top F_{tu}\delta y_u}_{E_{\text{cross}}}\\
&+\underbrace{\frac12\sum_t
\delta y_t^\top(F_{tt}-2s_tI)\delta y_t}_{E_{\text{anis}}}\\
&+o(\|\Delta\hat W\|_F^2).
\end{aligned}
\end{equation}

The approximation errors arise from cross-state Fisher interactions, directional anisotropy within individual Fisher blocks, and higher-order effects beyond the local second-order Taylor approximation. For simplicity, we omit the Monte Carlo estimation error of $s_t$ here.

\subsection{Optimality of Sensitivity-Weighted SVD}
\label{app:svd_optimality}

We show that \cref{eq:recons_2} admits an optimal rank-$r$ solution via SVD. Let $C=X_{\mathrm{on}}^\top D X_{\mathrm{on}}=S^\top S$, where $S$ is the Cholesky factor of the weighted covariance $C$. The objective can be rewritten as

\begin{equation}
\begin{aligned}
&\|D^{1/2}X_{\mathrm{on}}(W-\hat W)\|_F^2\\
&\qquad=\|S(W-\hat W)\|_F^2
=\|SW-S\hat W\|_F^2.
\end{aligned}
\end{equation}
The first equality follows from the trace representation of the Frobenius norm.

Assuming $S$ is invertible with damping factor, let $Z=S\hat W$, which preserves the rank constraint. By the Eckart--Young--Mirsky theorem, the optimal rank-$r$ approximation to $SW=U\Sigma V^\top$ is $Z^\star=U_r\Sigma_rV_r^\top$. Therefore,

\begin{equation}
\hat W^\star
= S^{-1}U_r\Sigma_rV_r^\top
= \sum_{i=1}^{r}\Delta W_i,
\end{equation}
where $\Delta W_i=S^{-1}u_i\sigma_i v_i^\top$ denotes the $i$-th singular component. Retaining the leading $r$ components with the largest singular values minimizes this objective for a fixed rank.

Thus, solving this surrogate objective yields a set of candidate singular components, whose squared singular values quantify their contributions to the surrogate error. Our global rank allocation then directly estimates the second-order impact of removing each component on the policy-KL divergence and uses these importance scores to determine which components to retain under a global budget. Since the surrogate error relies on simplifying approximations, the squared singular values and the policy-KL importance scores are not necessarily equivalent.
\section{Calibration and Evaluation Protocols}
\label{app:protocols}

\paragraph{Calibration Protocol.}
We use 256 fixed samples from the ScienceQA train split for calibration. For each model, we generate its own on-policy reasoning traces using greedy decoding, with a maximum of 512 generated tokens per sample. We collect all activations processed by each KV projection, including system prompts, multimodal input contexts, and generated reasoning tokens. For Fisher estimation, we independently sample next-token actions from the dense model's predictive distributions at all generation positions, using $K=8$ Monte Carlo samples by default, with a fixed random seed of 0. 

The calibration overhead primarily comes from generating on-policy reasoning traces for 256 examples and performing $K=8$ backward passes, taking approximately 30 minutes for a 7--8B model on a single NVIDIA L40S GPU.

\paragraph{Evaluation Protocol.}
We evaluate on the test split of ScienceQA (2017 samples), the testmini split of MathVista (1000 samples), and the validation split of MMMU (900 samples). All models use greedy decoding with a maximum of 4,096 generated tokens. We adopt the answer extraction and scoring procedures of lmms-eval~\cite{zhang2025lmmseval}, which extract answers from the final \textless answer\textgreater...\textless/answer\textgreater \ tags when available and apply the benchmark-specific fallback procedures otherwise. The same prompting protocol is used for calibration and evaluation.

\paragraph{Reasoning Prompts.}
All models share the following system prompt, with minor modifications to the default reasoning prompt of lmms-eval:

\begin{lstlisting}[
    basicstyle=\ttfamily\footnotesize,
    breaklines=true,
    breakindent=0pt,
    breakautoindent=false,
    columns=fullflexible,
    frame=single
]
You are a helpful assistant. When the user asks a question, your response must include two parts: first, the reasoning process enclosed in <think>...</think> tags, then the final answer enclosed in <answer>...</answer> tags. For multiple-choice questions, put only the option letter inside <answer>, for example: <answer>B</answer>. Otherwise, provide a clear, concise response within <answer> </answer> tags that directly addresses the question.
\end{lstlisting}

For Qwen3-VL-8B/32B-Thinking, we additionally append the following instruction to the user prompt:

\begin{lstlisting}[
    basicstyle=\ttfamily\footnotesize,
    breaklines=true,
    breakindent=0pt,
    breakautoindent=false,
    columns=fullflexible,
    frame=single
]
Reason concisely in a few sentences. After </think>, directly output the requested <answer>...</answer> with no additional explanation.
\end{lstlisting}
\section{Additional Results and Analysis}
\label{app:results}

\subsection{Alternative Baseline Calibration}
\label{app:baseline_calibration}
In our main experiments, we adopt a \textbf{example-matched calibration} protocol, where all methods use the same 256 ScienceQA train split examples. 
To further isolate the effects of calibration token count and reasoning-state coverage, we evaluate representative baselines under two additional protocols for collecting calibration activations: (a) \textbf{Token-matched calibration}, which adds more input-context examples until the total calibration token count matches our on-policy calibration; and (b) \textbf{GT-solution calibration}, which augments the original input contexts with ground-truth solution traces from the same 256 examples. 
For reference, the original 256 ScienceQA examples contain 171.7K tokens with input-only calibration, 190.9K tokens with GT solutions, and 200.3K tokens with on-policy reasoning traces. 

As shown in \cref{tab:app:baseline_calibration}, simply increasing the number of calibration tokens does not yield consistent improvement, while incorporating ground-truth solutions provides moderate gains. Nevertheless, our method consistently outperforms these enhanced baselines, highlighting the importance of on-policy calibration beyond increased token coverage. The component ablation in \cref{tab:ablation} and visualization analysis in \cref{fig:error_vis} further support that aligning calibration with on-policy reasoning states contributes the most to preserving reasoning performance.

\begin{table}[h]
\centering
\caption{Alternative calibration protocols on
Qwen2.5-VL-7B at 50\% rank ratio.
$\dagger$ denotes token-matched input-context calibration;
$\ddagger$ denotes calibration with ground-truth solutions.}
\label{tab:app:baseline_calibration}
\small
\begin{adjustbox}{max width=\linewidth}
\begin{tabular*}{\linewidth}
{@{\extracolsep{\fill}}lcccc@{}}
\toprule
\textbf{Method} & \textbf{SciQA} &
\textbf{MathV} & \textbf{MMMU} & \textbf{Avg.} \\
\midrule
SVD-LLM                    & 82.75 & 54.6 & 37.89 & 58.41 \\
SVD-LLM$^{\dagger}$        & 81.36 & 54.3 & 39.22 & 58.29 \\
SVD-LLM$^{\ddagger}$       & 83.74 & 57.2 & 40.89 & 60.61 \\
\midrule
Palu                       & 74.17 & 55.1 & 43.77 & 57.68 \\
Palu$^{\dagger}$           & 75.76 & 53.7 & 42.56 & 57.34 \\
Palu$^{\ddagger}$          & 84.86 & 58.5 & 46.11 & 63.16 \\
\midrule
\textbf{Ours}           & \textbf{87.40} & \textbf{63.2}
& \textbf{49.11} & \textbf{66.57} \\
\bottomrule
\end{tabular*}\
\end{adjustbox}
\end{table}

\subsection{Monte Carlo Estimation Stability}
\label{app:mc_stability}
We examine the stability of our Monte Carlo estimators with $K\in\{1,2,4,8,16\}$, using an independent estimate with $K_{\mathrm{ref}}=64$ as reference. For sensitivity-weighted covariance, we compute the relative Frobenius error for each KV projection and average across all projections. For singular-component importance, we measure the global Spearman rank correlation across all components, with the SVD bases held fixed. Results are averaged over three random seeds with standard deviation.

\begin{figure}[t]
    \centering
    \includegraphics[width=\linewidth]{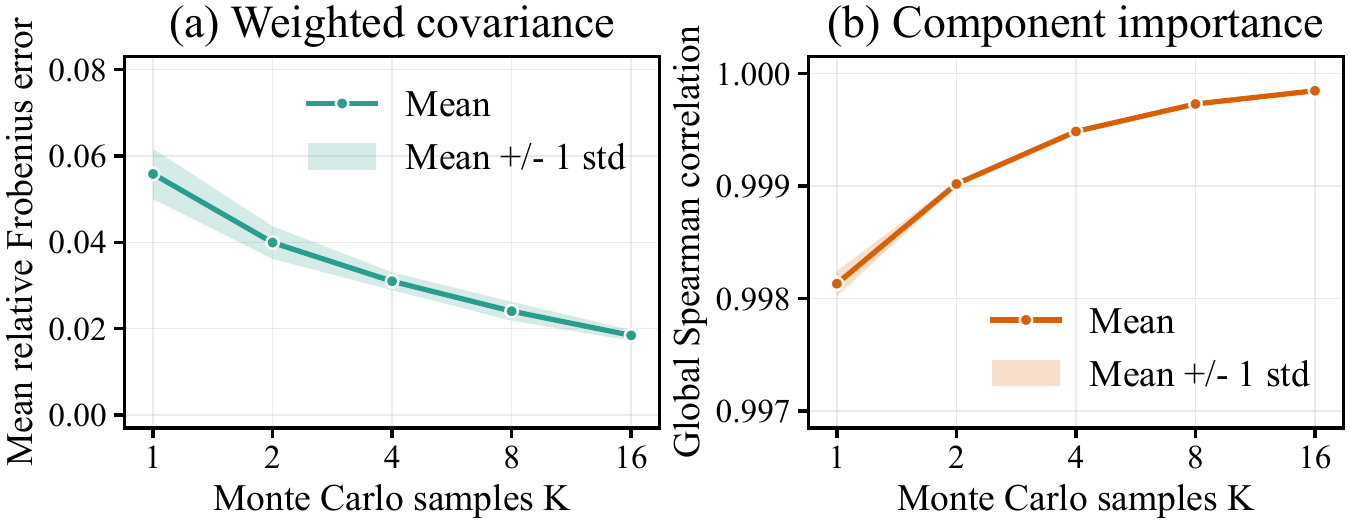}
    \caption{Monte Carlo estimation stability on Qwen2.5-VL-7B-Instruct. (a) Mean relative Frobenius error of sensitivity-weighted covariance matrices. (b) Global Spearman correlation of singular-component importance scores. Results show mean $\pm$ standard deviation over three random seeds, using $K_{\mathrm{ref}}=64$ as reference.}
    \label{fig:mc_stability}
\end{figure}

As shown in \cref{fig:mc_stability}, the covariance estimation error decreases steadily as $K$ increases, reaching approximately 2.4\% at the default $K=8$. Meanwhile, component importance rankings remain highly consistent, with Spearman correlations exceeding 0.998 even at $K=1$ and approaching 1 as $K$ increases. These results demonstrate the stability of our Monte Carlo estimators and support the use of a modest sample count ($K=8$) in practice.

\subsection{Generation Length on Thinking Models}

We analyze the generation lengths and truncation rates of Qwen3-VL-8B-Thinking in \cref{tab:app:length}. Unlike general-purpose VLMs such as Qwen2.5-VL-7B-Instruct, we observe that errors on difficult questions in this thinking-tuned model often manifest as \textbf{repetitive reasoning without producing a certain answer}, rather than terminating with an incorrect answer~\cite{pipis2026wait}. Non-truncated generations are typically much shorter (e.g., only avg. 680 tokens for the dense model on MMMU), and increasing the token limit to 8,192 yields little improvements, suggesting that such failures are not merely caused by an insufficient token budget.

KV compression can further exacerbate such behavior. 
On MMMU at 50\% rank ratio, SVD-LLM increases the average generation length from 1,280 to 1,943 tokens, with the truncation rate rising from 17.56\% to 33.44\%. In contrast, TraceKV limits the average length to 1,559 tokens and the truncation rate to 23.22\%, substantially mitigating this issue.
Similar trends hold across other benchmarks and quantized settings. These findings suggest that our method better preserves the original reasoning behavior, while mitigating the additional per-sample decoding overhead associated with prolonged reasoning failures.

\begin{table}[t]
\centering
\caption{Generation length and truncation analysis on
Qwen3-VL-8B-Thinking at 50\% rank ratio.
Each entry reports average generated tokens / truncation rate
under a 4,096-token limit.}
\label{tab:app:length}
\small
\begin{adjustbox}{max width=\linewidth}
\begin{tabular}{@{}lccc@{}}
\toprule
\textbf{Method} & \textbf{ScienceQA} & \textbf{MathVista} & \textbf{MMMU} \\
\midrule
Dense
& 214 / 1.59\% & 700 / 6.50\% & 1,280 / 17.56\% \\
\midrule
SVD-LLM
& 443 / 10.91\% & 1,343 / 18.00\% & 1,943 / 33.44\% \\
Palu
& 396 / 7.59\% & 1,456 / 20.50\% & 1,871 / 30.89\% \\
\textbf{Ours}
& 260 / 2.93\% & 1,121 / 10.40\% & 1,559 / 23.22\% \\
\midrule
Palu (KV4)
& 407 / 7.93\% & 1,463 / 20.80\% & 1,959 / 31.33\% \\
\textbf{Ours (KV4)}
& 276 / 3.07\% & 1,170 / 11.70\% & 1,637 / 25.67\% \\
\bottomrule
\end{tabular}
\end{adjustbox}
\end{table}

\subsection{Analysis on KV-State Sensitivity}

\begin{figure}[t]
    \centering
    \includegraphics[]{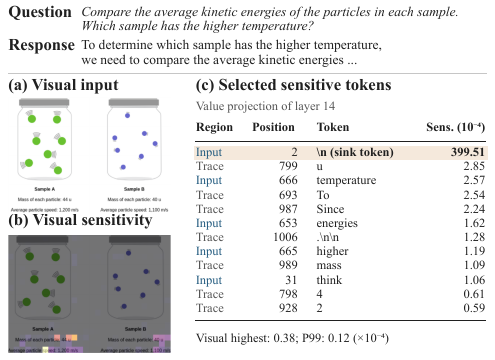}
    \caption{Visualization of KV-state sensitivity for a ScienceQA example on Qwen2.5-VL-7B-Instruct, measured at the value projection of layer 14.}
    \label{fig:app:sensitivity}
\end{figure}

\cref{fig:app:sensitivity} provides a qualitative visualization of policy-KL sensitivities of KV states. 
As shown, high sensitivity among visual tokens is concentrated in a few regions containing question-relevant numerical information, while most visual regions exhibit substantially lower sensitivity.
In contrast, textual tokens generally receive higher sensitivity scores, particularly reasoning-critical tokens in both the input context and generated reasoning trace.
Notably, the initial attention sink token exhibits exceptionally high sensitivity, consistent with prior findings on attention sinks~\cite{xiao2024streamingllm,hooper2024kvquant}. In practice, the sink token has an identical activation across different samples, and its KV state can theoretically be reconstructed with a single rank-one component, allowing the remaining rank budget to accommodate other tokens despite the sink token's disproportionately high sensitivity.

\subsection{Analysis on Rank Allocation}
\label{app:rank_alloc}

\begin{figure}[t]
    \centering
    \includegraphics[]{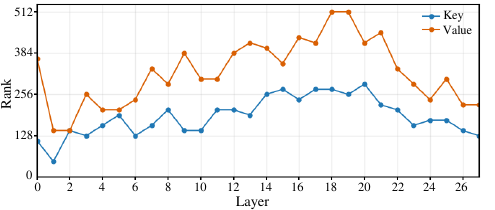}
    \caption{Visualization of global rank allocation on Qwen2.5-VL-7B-Instruct at 50\% rank retention.}
    \label{fig:app:rank_alloc}
\end{figure}

\cref{fig:app:rank_alloc} visualizes the layer-wise rank allocation for Qwen2.5-VL-7B-Instruct under a global rank ratio of 50\%. 
The allocated ranks exhibit substantial variation across both layers and projection types. In particular, value projections receive larger rank budgets than key projections in most layers, while both generally receive more capacity in the middle-to-late layers, especially around layers 16--21. These observations reveal the heterogeneous importance of singular components across KV projections, highlighting the need for importance-guided global rank allocation rather than uniform rank assignment.

\section{Additional Implementation Details}
\label{app:implementation}

\subsection{Quantization Integration}
\label{app:quantization}

To facilitate Hadamard rotation, we align the allocated ranks to multiples of 16. We apply an orthogonal Hadamard transform to the latent dimensions, which can be absorbed into the low-rank factors without additional inference overhead. Inspired by QSVD~\cite{wang2026qsvd}, we further redistribute the singular values between the two factors. Specifically, for each KV projection $W_m\approx A_mB_m$, we construct
\begin{equation}
\begin{aligned}
A_m &= S_m^{-1}U_{\mathcal I_m}
\Sigma_{\mathcal I_m}^{\alpha}H_m,\\
B_m &= H_m^\top
\Sigma_{\mathcal I_m}^{1-\alpha}
V_{\mathcal I_m}^\top,
\end{aligned}
\end{equation}
where $H_m$ is a normalized Hadamard matrix. 
Although this $\alpha$-redistribution leaves $A_mB_m$ largely unchanged in BF16 precision, the symmetric choice $\alpha=0.5$ is not necessarily optimal under latent quantization. Specifically, the down-projection factor $A_m$ controls the magnitude of latent activations, thereby affecting their quantization error, while the up-projection factor $B_m$ determines how this error is amplified during KV-state reconstruction.

Empirically, we find that $\alpha=0.7$ consistently reduces the quantization-induced normalized mean squared error (NMSE) of reconstructed KV states by 4--8\% relative to $\alpha=0.5$ across our evaluated models. We therefore fix $\alpha=0.7$ for all experiments without model-specific tuning. \cref{tab:app:quant_ablation} further examines the downstream effects of Hadamard rotation and singular-value redistribution under 4-bit and 3-bit latent quantization.

\begin{table}[t]
\centering
\caption{Ablation of Hadamard rotation and singular-value
redistribution on Qwen2.5-VL-7B-Instruct at 50\% rank
ratio, with 4-bit and 3-bit latent quantization.
The default configuration uses Hadamard rotation
and $\alpha=0.7$.}
\label{tab:app:quant_ablation}
\small
\begin{adjustbox}{max width=\linewidth}
\begin{tabular*}{\linewidth}
{@{\extracolsep{\fill}}llcccc@{}}
\toprule
\textbf{Bits} & \textbf{Variant} &
\textbf{SciQA} & \textbf{MathV} &
\textbf{MMMU} & \textbf{Avg.} \\
\midrule
\multirow{3}{*}{KV4}
 & Ours (default)      & 87.41 & 62.7 & 49.89 & 66.67 \\
 & w/o Hadamard        & 87.41 & 62.4 & 48.56 & 66.12 \\
 & $\alpha=0.5$        & 88.15 & 62.1 & 47.67 & 65.97 \\
\midrule
\multirow{3}{*}{KV3}
 & Ours (default)      & 87.36 & 62.5 & 47.56 & 65.80 \\
 & w/o Hadamard        & 86.07 & 61.8 & 48.56 & 65.47 \\
 & $\alpha=0.5$        & 87.11 & 61.6 & 46.67 & 65.13 \\
\bottomrule
\end{tabular*}
\end{adjustbox}
\end{table}

\subsection{Efficient Low-Rank Attention}
\label{app:attention_impl}

We implement a two-stage decode pipeline for Qwen2.5-VL-7B-Instruct that avoids materializing full historical K/V cache in global memory. A fused Triton QK kernel reconstructs key tiles on-chip, applies MRoPE using cached position IDs, and reuses each key tile across its associated query heads under grouped-query attention (GQA). Attention weights then aggregate values directly in the latent space, followed by the value up-projection, so value reconstruction is performed only after reduction over the sequence dimension. Both BF16 and INT4 caches follow this pipeline; the INT4 kernels unpack codes on-chip and factor affine quantization corrections out of the inner products. Prefill still uses standard attention with reconstructed K/V.

\begin{figure}[t]
    \centering
    \includegraphics[width=\linewidth]{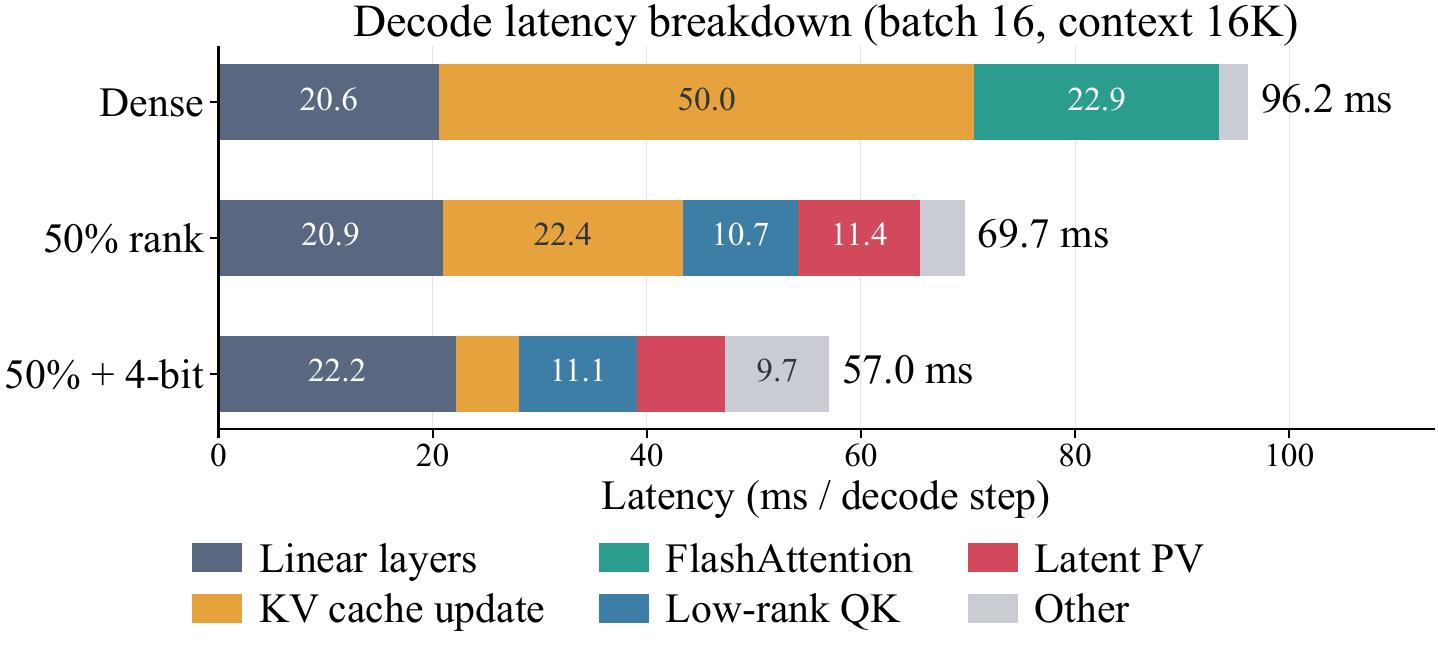}
    \caption{Per-step decoding latency breakdown on Qwen2.5-VL-7B-Instruct with batch size 16 and 16K context length. Reported latencies may differ slightly from end-to-end measurements due to the overhead of kernel-level profiling.}
    \label{fig:app:latency}
\end{figure}

\cref{fig:app:latency} presents the decoding latency breakdown under different compression settings. Under the default Hugging Face implementation, TraceKV substantially reduces KV-cache update overhead by storing compact latent cache, with 4-bit quantization providing further reductions in memory traffic. Despite requiring KV reconstruction for attention computation, our optimized low-rank attention remains efficient, with the combined QK and PV kernel latency comparable to or lower than dense FlashAttention-2. 

\section{Limitations and Discussion}

\paragraph{Compression Objective Alignment.}
TraceKV is guided by a unified policy-preserving principle that minimizes policy-KL divergence over reasoning traces generated by the dense model, following the widely adopted objective of knowledge distillation. However, this objective does not necessarily guarantee optimal performance of the compressed model, which encounters its own autoregressive states during inference rather than those of the dense model. Such distribution mismatch may lead to accumulated errors over long reasoning trajectories. Exploring compression objectives that explicitly account for compressed-model traces is a promising direction for future work.

\paragraph{Calibration Overhead and Generalization.}
TraceKV introduces additional calibration overhead from on-policy trace generation and Monte Carlo estimation. Moreover, although our experiments demonstrate generalization across several downstream benchmarks, calibration still relies on a limited set of task-specific examples and their model-generated reasoning traces. Whether the resulting compression strategy generalizes to broader domains or other tasks involving long-form generation beyond reasoning remains an interesting direction for future research.

\paragraph{Comparison with Online Low-Rank Methods.} 
Another line of work explores online low-rank KV compression by directly factorizing prefill KV cache matrices on-the-fly~\cite{sun2025shadowkv,chang2026xkv}. However, the resulting low-rank bases may not adequately cover newly generated states during reasoning, while updating the bases and historical caches online remains expensive~\cite{zhao2026iskv}. 
In contrast, TraceKV calibrates fixed low-rank projection using on-policy reasoning traces for compressing both input and generated KV states without additional runtime cost.

\end{document}